\pdfoutput=1

\documentclass[11pt]{article}

\usepackage[final]{acl}

\usepackage{times}
\usepackage{latexsym}

\usepackage{amsmath}
\usepackage{amsfonts}
\usepackage{amssymb}
\usepackage{algorithm}
\usepackage{algpseudocode}
\usepackage{cleveref}
\usepackage{graphicx}

\usepackage{natbib}

\usepackage{caption}
\usepackage{hyperref}

\usepackage{booktabs}
\usepackage{hyperref}
\usepackage{multirow}
\usepackage{pbox}  %
\usepackage{enumitem} %
\usepackage{array} %
\newcolumntype{L}[1]{>{\raggedright\arraybackslash}p{#1}} %
\newcolumntype{C}[1]{>{\centering\arraybackslash}p{#1}}   %
\usepackage{xcolor}     %
\usepackage{array}      %

\newcolumntype{H}[1]{>{\columncolor{#1}}c}

\usepackage[T1]{fontenc}

\usepackage[utf8]{inputenc}

\usepackage{microtype}

\usepackage{inconsolata}

\usepackage{graphicx}

\title{Learning Natural Language Constraints for Safe Reinforcement Learning of Language Agents}

\author{
  Jaymari Chua \\
  CSIRO's Data61 and UNSW\\
  Sydney, NSW, Australia \\
  \And
  Chen Wang \\
  CSIRO's Data61\\
  Sydney, NSW, Australia \\
  \texttt{\{jace.chua, chen.wang, lina.yao\}@data61.csiro.au}
  \And
  Lina Yao \\
  CSIRO's Data61 and UNSW\\
  Sydney, NSW, Australia \\
}

\begin{document}
\maketitle
\begin{abstract}
Generalizable alignment is a core challenge for deploying Large Language Models (LLMs) safely in real-world NLP applications. Current alignment methods, including Reinforcement Learning from Human Feedback (RLHF), often fail to guarantee constraint satisfaction outside their training distribution due to their reliance on implicit, post-hoc preferences. Inspired by a paradigm shift to first curate data before tuning, we introduce a new framework for safe language alignment that learns natural language constraints from positive and negative demonstrations as a primary step. From inferring both a task-specific reward function and latent constraint functions, our approach fosters adaptation to novel safety requirements and robust generalization under domain shifts and adversarial inputs. We formalize the framework within a Constrained Markov Decision Process (CMDP) and validate it via a text-based navigation environment, demonstrating safe adaptation to changing danger zones. Our experiments show fewer violations upon domain shift when following a safe navigation path, and we achieve zero violations by applying learned constraints to a distilled BERT model as a fine-tuning technique. This work offers a promising path toward building safety-critical and more generalizable LLMs for practical NLP settings.
\end{abstract}

\section{Introduction} \label{sec:introduction}

Large language models (LLMs) are increasingly entrusted with high-stakes decisions in domains ranging from legal advisory to healthcare triage \cite{wang2021evidence, zhang2023survey}, where open-ended deployments expose critical safety gaps under domain shifts \cite{yang2021revisiting, moskovitz2023fragility}. Ensuring LLMs remain reliable in unpredictable contexts is paramount for averting harmful or misguided recommendations \cite{bai2022training, casper2023open}. As these models improve and learn new capabilities, the challenge shifts from straightforward compliance in known conditions to achieving alignment requirements that safeguards against edge case mistakes and risks arising from diverse, evolving environments \cite{gao2022scaling, rottger2024inference}.

Particularly for LLMs being used as base or foundation models, the current alignment training methods struggle to maintain safe and reliable behavior when faced with adversarial prompts or subtle environmental variations. Despite broad adoption, Reinforcement Learning from Human Feedback (RLHF) often lacks deep causal grounding \cite{di2022goal, hadfield2017inverse} and depends heavily on post-hoc reward adjustments \cite{stiennon2020learning, ouyang2022training}. This reactive design can invite reward overfitting \cite{gao2022scaling}, leading to degenerate policies that narrowly exploit preference models \cite{rottger2024inference} and underperform out of distribution \cite{saleh2020resource, casper2023open}. While RLHF can yield surface-level compliance, it offers no guarantees of reliable behavior when contexts shift or when adversarial prompts appear \cite{moskovitz2023fragility, jin2020bert}. This can lead to degenerate behaviors and poor performance when the model encounters situations outside of its training distribution \cite{saleh2020resource, casper2023open}. In essence, RLHF struggles to enforce explicit safety rules, particularly those that can be concisely expressed in natural language. We posit that preference learning alignment has synergy with a proactive safe RL paradigm, one that formalizes and minimizes high-risk actions rather than relying on human feedback alone to retroactively shape model outputs \cite{yang2021revisiting, bai2022training}.

Our work is a framework for natural language constraint learning from text demonstrations within safe reinforcement learning. Building upon the foundational work on inverse reinforcement learning with learned constraints \cite{hadfield2017inverse, arora2021survey}, our approach leverages Constrained Markov Decision Processes (CMDPs) \cite{achiam2017constrained} and risk-averse reinforcement learning \cite{chow2018lyapunov} to infer both a task-specific reward function and latent safety constraints, expressed in natural language. These are learned initially from positive and negative demonstrations, and then further refined through interaction with the environment. While prior work has explored interpreting pre-defined natural language constraints \cite{lou2024safe, feng2024natural} or modifying reward functions for classification tasks \cite{liao-etal-2024-enhancing}, our framework extends inverse reinforcement learning to learn these constraints, promoting adaptation to novel safety requirements and robust generalization across diverse NLP tasks and environments.

Our key contributions are threefold: (1) extending inverse reinforcement learning to learn natural language safety constraints from a combination of demonstrations and environmental interaction; (2) formalizing generalizable safety alignment as a CMDP and as a constrained inverse reinforcement learning technique, to infer both reward and constraint functions in natural language; and (3) empirically demonstrating, through a proof-of-concept experiment in a text-based navigation environment, improved robustness to distributional shifts and adversarial prompts compared to standard RLHF, achieved by proactively minimizing high-risk decisions.

The remainder of this paper details our framework. \Cref{sec:background} situates our approach within related alignment and safe RL work. \Cref{sec:framework} is our natural language constraint learning framework, including the problem formulation and the method for adapting constraint-learning inverse reinforcement learning for inferring said constraints from text demonstrations. \Cref{sec:experiment} is our experiment to demonstrate feasibility of the framework, \Cref{sec:open} discusses implications for generalizable alignment and identifies open challenges in natural language RL. \Cref{sec:conclusion,sec:limitations} conclude and acknowledge limitations.

\section{Background} \label{sec:background}

\paragraph{Generalizable Alignment} Large Language Models (LLMs) are base models that drive the decision-making process of language agents. Superalignment \cite{burns2023weak, ngo2022alignment}, is the open problem of ensuring that AI systems far exceeding human intelligence remain aligned with human intent across all domains. Given the increasing deployment and wide use of large language models (LLMs) in high-stakes decision-making, even before the advent of such \emph{superhuman AI}, robust alignment techniques are urgently needed. Existing techniques are provably insufficient in guaranteeing robustness to all possible inputs and generalization across all potential domain shifts.

\paragraph{Large Language Model (LLM) Training} Although standard LLM training incorporates elements of robustness and generalization across its stages, these strategies alone may not suffice to meet the exacting demands of Generalizable alignment. LLM development begins with pre-training on massive text corpora, yielding foundational models such as BERT, GPT-2, and GPT-3, and scaling to architectures like PaLM, GLaM, and Chinchilla \cite{devlin-etal-2019-bert, brown2020language, radford2019language, chowdhery2023palm, du2022glam, hoffmann2022training}. While this large-scale pre-training confers broad linguistic and world knowledge, it is insufficient for achieving the stable performance under adversarial or shifting conditions, i.e. robustness, and the ability to succeed on previously unseen tasks, i.e. generalization, that are required for generalizable alignment. To address these gaps in the next phase, fine-tuning applies a range of methods. Supervised learning \cite{rajpurkar-etal-2016-squad, socher-etal-2013-recursive} and domain adaptation \cite{gururangan-etal-2020-dont} extend the model’s applicability to new tasks and contexts, thereby improving generalization. Instruction tuning (e.g., FLAN, T0) \cite{wei2021finetuned, sanh2021multitask} likewise enhances generalization by tuning the model more effectively with task instructions. Additionally, parameter-efficient approaches such as LoRA \cite{hu2021lora} refine model performance without needing full model retraining, maintaining strong generalization while reducing computational overhead. In contrast, adversarial training \cite{goodfellow2014explaining, miyato2018virtual} improves robustness by exposing models to harder or perturbed examples, boosting resilience to input variations. Multilingual and multi-task setups in BLOOM \cite{conneau-etal-2020-unsupervised, xue-etal-2021-mt5, le2023bloom}, further reinforce both generalization and robustness by training on diverse linguistic contexts. Despite performance gains in adaptability, aligning model behavior with human values motivates a dedicated alignment training phase, centered on Reinforcement Learning from Human Feedback (RLHF) \cite{christiano2017deep, ouyang2022training}, with ongoing investigations into alternatives such as Constitutional AI or Reinforcement Learning from AI Feedback (RLAIF) and Direct Preference Optimization (DPO) \cite{bai2022constitutional, rafailov2024direct}. However, preference reinforcement learning strategies may exhibit failure modes that undermine their utility for superalignment objectives or, at a minimum, their effectiveness in further fine-tuning for diverse domain adaptations.

\subsection{Mechanistic Failures of RLHF in Achieving Robust Generalization} \label{subsec:rlhf}
Across many studies, RLHF has yielded substantial gains in aligning model behavior with user preferences. The RLHF paradigm involves fine-tuning a pretrained model through a cyclical process of human feedback collection, reward model training, and policy optimization \cite{christiano2017deep, ziegler2019fine, ouyang2022training}.  Nonetheless, several mechanistic failures hinder its ability to achieve deep, reliable generation across novel conversations, unseen texts, and complex reasoning tasks. First, reward models, often constructed from limited human annotations, can misattribute high reward to superficial linguistic features (e.g., tone, formality, length) rather than capturing the true intent behind human judgments. This causal misattribution, even with regularizers like KL-divergence, can lead to reward hacking and mode collapse during overoptimization \cite{stiennon2020learning, ouyang2022training, pan2022effects, gao2022scaling, glaese2022improved, casper2023open}. Second, RLHF policies are prone to reward model drift and exposure bias, particularly with out-of-distribution inputs or long-horizon tasks, leading to unsafe or incoherent responses \cite{perez2022red, kirk2023understanding, ramamurthy2023much}. Finally, concerning generalization, the fine-tuning process in RLHF can create rigid prompt-response mappings, limiting compositional generalization and multi-hop reasoning which is crucial for tasks requiring diverse knowledge integration \cite{lampinen2022can, dziri2023faith, casper2023open}.

\subsection{Path to Generalizable Alignment: Safe RL} \label{subsec:safe-rl}
Safe RL offers a principled approach to LLM alignment, shifting from implicit alignment via feedback to explicit alignment through constrained optimization and risk management. A key development is Safe RLHF, which incorporates human feedback within a Constrained Markov Decision Process (CMDP) framework \cite{ray2019benchmarking, yang2021worry}. These algorithms fine-tune the LLM to maximize a reward model representing helpfulness while simultaneously ensuring that a learned safety metric remains below a predefined threshold \cite{ray2019benchmarking, yang2021worry}. Empirical results demonstrate that this approach can mitigate harmful outputs more effectively than standard RLHF, without significant performance degradation on helpfulness \cite{ray2019benchmarking, dai2023safe}. Decoupling helpfulness and harmlessness into separate objectives, Safe RLHF avoids the trade-offs inherent in a single reward function \cite{ray2019benchmarking, ma2023following}. This results in a policy that internalizes constraints against unsafe behavior, providing a stronger safety guarantee than policies that simply try to avoid low-reward outputs during training. Safe RL directly addresses several failure modes of standard RLHF. Reward hacking is mitigated because the training algorithm penalizes or deems infeasible any attempt to maximize reward by violating safety constraints \cite{chow2018lyapunov, achiam2017constrained, ray2019benchmarking}. Safe RL can also reduce sycophancy by incorporating truthfulness or consistency as constraints or additional reward signals, rather than solely optimizing for human approval \cite{perez2022discovering, ouyang2022training}. Furthermore, adversarial prompts and jailbreaks are less effective when the model's policy has been trained to avoid generating forbidden content altogether, due to the imposed constraints \cite{ray2019benchmarking, yang2021worry, wei2023jailbreak}. In essence, Safe RL instills a form of robust rule-following within the model's policy, whereas RLHF's safeguards can be more easily circumvented outside the narrow distribution of training data \cite{ray2019benchmarking, bai2022constitutional}. Safe RL incorporates risk awareness, among the safety requirements of generalizable language models, where even infrequent dangerous outputs are unacceptable \cite{bostrom2014superintelligence, russell2019human}.

\subsection{Enhanced Generalization: IRL}
\label{subsec:irl} While Safe RL in the previous section, \cref{subsec:safe-rl}, enforces safety constraints, it still depends on explicitly defining human preferences as rewards. Inverse Reinforcement Learning (IRL) on the other hand, infers latent reward functions directly from expert demonstrations \cite{ng2000algorithms, abbeel2004apprenticeship}, bypassing these limitations. As such, IRL addresses others of the RLHF's limitations discussed in \cref{subsec:rlhf}, specifically its reliance on potentially noisy or superficial human feedback, offering even more improved performance across domains. IRL in modern research extends to high-dimensional settings and incorporates adversarial techniques \cite{ziebart2008maximum, wulfmeier2015maximum, ho2016generative}. More recent work adapts IRL to language, exploring natural language explanations \cite{li2023inferring, yu2024language, zhang2024aligning}, mitigating LLM-specific failure modes \cite{kent2023parametrizing, zhang2024rl}, and combining IRL with preference learning \cite{xu2023preference, zhang2024aligning}. As such, IRL uncovers underlying reward functions and promotes generalization to novel inputs and complex reasoning, avoiding the rigid mappings of RLHF \cite{syed2008game, levine2011nonlinear}.

\paragraph{Similar Work} Concurrent with our work, \citet{sun2024inverse} explore LLM alignment through demonstration data in their Inverse-RLignment framework, focusing on learning a standard reward function but to compared to ours, theirs is without explicitly modeling safety constraints. In contrast, \citet{lou2024safe} rely on pre-trained LMs to interpret predefined natural language constraints, whereas our own framework learns these constraints directly from demonstrations, enabling adaptation to new safety concerns. Our approach also extends prior inverse constrained RL methods \cite{xu2023preference} to high-dimensional language models under adversarial settings, and the first one integrating IRL with safe RL frameworks fundamentally as CMDPs \cite{altman1999constrained} for robust constraint enforcement. Another similar work that inspired our frameworkd is NLRL; \citet{feng2024natural} introduced Natural Language Reinforcement Learning (NLRL) to represent RL concepts entirely in natural language, they neither address safety constraints nor employ IRL. Finally, \citet{liao-etal-2024-enhancing} propose Reinforcement Learning framework with Label-sensitive Reward (RLLR) to improve classification tasks in RLHF for natural language understanding, whereas our natural language constraint learning framework tackles sequential decision-making by learning separate constraint functions that govern acceptable behavior, irrespective of the task reward; and it is our formal framework that makes use of the synergy of IRL with safe RL, as our framework offers a flexible, relatively more scalable approach to reliably aligning language-driven agents in dynamic environments.

\section{Learning Natural Language Constraints: A Framework} \label{sec:framework}

\subsection{Preliminaries}

\subsubsection{Safe RL in NLP: A Constrained MDP Framework}

Safe reinforcement learning is based on a \emph{Constrained Markov Decision Process (CMDP)} \cite{altman1999constrained}, and essentially can be used for language modeling by defining $(\mathcal{S}, \mathcal{A}, T, R, C, \gamma)$, where $\mathcal{S}$ is the (textual) state space, $\mathcal{A}$ the action space (e.g., text outputs), $T$ the transition function, $R$ the reward, $C$ a cost for unsafe behavior, and $\gamma$ the discount factor. The objective is to maximize a policy $\pi$ satisfying:
\[
\begin{aligned}
\max_{\pi}\quad & \mathbb{E}_\pi\!\left[\sum_{t=0}^{\infty}\gamma^t R(s_t,a_t)\right], \\
\text{s.t.}\quad & \mathbb{E}_\pi\!\left[\sum_{t=0}^{\infty}\gamma^t C(s_t,a_t)\right] \le H.
\end{aligned}
\]
wherefore safety, $C(s,a)$ can capture constraints such as for preserving privacy, and extend this framework by incorporating \emph{free-form text constraints}: from defining a constraint space $\mathcal{X}$ of natural language rules and a mapping $M: \mathcal{X} \to C$ that translates a rule (e.g., “Do not reveal private data”) into a cost function, thereby enabling the agent to receive safety instructions in natural language and incorporate them directly into model training \cite{yang2021safe, lou2024safe}.

\subsection{Framework Overview}

This paper introduces a novel framework for developing safer large language models (LLMs) and creates a synergy of approaches which we now call \emph{natural language constraint learning}.

As the fundamental limitations of Reinforcement Learning from Human Feedback (RLHF) and existing Safe RL methods as detailed in \cref{sec:background} drive our research agenda: failure modes to reward hacking, brittleness to distribution shift, reliance on implicit constraints, and lack of transparency necessitate a paradigm shift. Existing Safe RL often assumes perfectly known, a priori safety constraints, which is an unrealistic assumption in complex, real-world scenarios. In generalization performance, RLHF comes with an alignment performance cost.

Our framework is grounded on CMDPs and address these limitations through three interconnected ideas: our own Constraint Learning via Inverse Reinforcement Learning (CLIRL) \cref{subsec:cl-irl}, Constraint Aware Policy Optimization (CAPO) \cref{subsec:capo}, and Conditional Value at Risk (CVaR) \cref{subsec:cvar}. CLIRL simultaneously learns a reward function (for task performance) and constraint functions (for safety) from a separated class of positive and negative demonstrations, a departure from standard Inverse Reinforcement Learning. CAPO utilizes the learned constraints to ensure that policy updates remain within a safe region. We model domain shifts and adversarial inputs by incorporating stochastic environment transitions and employ CVaR minimization to satisfy constraints.

\subsection{Problem Formulation: The Constrained Markov Decision Process (CMDP)}

We formalize safe language generation as a CMDP, $(\mathcal{S}, \mathcal{A}, T, R, \mathcal{C}, \gamma, H)$. The state space, $\mathcal{S}$, represents textual context: dialogue history, prompts, and retrieved knowledge. Each $s \in \mathcal{S}$ is a token sequence. The action space, $\mathcal{A}$, encompasses all possible next tokens; $a \in \mathcal{A}$ appends a token.

The transition function, $T(s' | s, a, \theta)$, gives the probability of reaching $s'$ from $s$ given $a$ and domain parameter $\theta \in \Theta$. This stochasticity models domain shifts and adversarial perturbations. The reward, $R(s, a)$, signifies "helpfulness". We learn $R$ via CLIRL \cref{subsec:cl-irl}.

The constraint set, $\mathcal{C}$, has $K$ functions, $C_k(s, a)$, $k = 1, ..., K$, each quantifying the cost of violating a safety constraint (e.g., toxicity). These are also learned. $\gamma \in [0, 1]$ is the discount factor. $H = [H_1, ..., H_K]$ is the constraint threshold vector; $H_k$ is the maximum cumulative discounted cost for $C_k$.

\subsection{Constraint Learning Inverse Reinforcement Learning (CLIRL)} \label{subsec:cl-irl}

The core innovation of our framework is Constraint Learning Inverse Reinforcment Learning (CLIRL), changing IRL to learn rewards and constraints. We use positive demonstrations, $D_{pos}$ ($\{\tau_i^+\}$ of desirable behavior), and negative demonstrations, $D_{neg}$ ($\{\tau_j^-\}$ of undesirable behavior), and details of the objective is detailed in \cref{subsec:cl-irl-obj}.

After policy learning our method discovers safety constraints, not manual specifications. Negative demonstrations are key. For example, in a dialogue setting, a negative demonstration might be a conversation turn where the LLM generates a toxic response, reveals private information, or provides a factually incorrect answer. In a text-based game, a negative demonstration could be a sequence of actions that leads to a game-over state due to violating a safety rule (e.g., drinking a poisonous potion or walking into a bottomless pit).

\begin{table}[h!]
\centering
\caption{Positive and Negative Demonstrations}
\label{tab:examples}
\small
\begin{tabular}{@{}L{0.45\columnwidth}L{0.45\columnwidth}@{}}
\toprule
\textbf{Positive Demonstration (Dialogue)} & \textbf{Negative Demonstration (Dialogue)} \\
\midrule
User: What's the capital of France? & User: What's the capital of France? \\
LLM: The capital of France is Paris. & LLM: The capital of France is Berlin. You idiot! \\
\midrule
\textbf{Positive Demonstration (Text Game)} & \textbf{Negative Demonstration (Text Game)} \\
\midrule
\texttt{> go north} & \texttt{> drink poison potion} \\
You are in a serene place. & You feel a burning sensation... You have died! \\
\texttt{> take key} &  \\
You pick up the key. &  \\
\bottomrule
\end{tabular}
\end{table}

In traditional NLP settings i.e. toxicity, a toxicity constraint function can be learnt, $C_{toxicity}(s, a)$, might be implemented as a neural network that takes the current state (dialogue history) $s$ and the proposed next action (word) $a$ as input and outputs a score representing the likelihood of the resulting text being toxic. This network could be pre-trained on a large dataset of toxic and non-toxic text, or it could be fine-tuned during the CLIRL process.

In the extended environments and adapted use cases for language agents, another constraint, $C_{factual}(s,a)$, could measure the consistency of the generated text with a world understanding knowledge base. For instance, the domain may change as $\theta_1$ might represent standard, grammatically correct English text.  $\theta_2$ could represent text with common misspellings and grammatical errors. $\theta_3$ might represent text with adversarial perturbations specifically designed to trigger toxic outputs. By training on a distribution over these different text-based representations of worlds as domains encoded as $\theta$ values, we encourage the model to be robust to a wide range of input variations.

\subsection{Constraint-Aware Policy Optimization} \label{subsec:capo}

After learning $R_{\theta}$ and $C_{k, \phi_k}$ via CLIRL, we train a policy $\pi_{\psi}(a|s)$ (parameterized by $\psi$) using Constraint-Aware Policy Optimization (CAPO), a modified CPO. CAPO's objective:

\begin{multline}
J_{CAPO}(\psi) = \mathbb{E}_{\tau \sim \pi_{\psi}}\left[\sum_{t} \gamma^t R_{\theta}(s_t, a_t)\right] \\ 
- \sum_{k=1}^K \beta_k \mathbb{E}_{\tau \sim \pi_{\psi}}\left[\sum_{t} \gamma^t C_{k, \phi_k}(s_t, a_t)\right]
\end{multline}
where $\beta_k$ are dynamic Lagrange multipliers. CAPO uses trust region optimization, ensuring each update improves reward and satisfies constraints, preventing reward exploitation.
\begin{algorithm}[ht!]
\caption{Natural Language Constraint Learning Framework, Applied}
\begin{algorithmic}[1]
\State \textbf{Input:} $D_{pos}$, $D_{neg}$, $H$
\State \textbf{Output:} $\pi_{\psi}$, $R_{\theta}$, $C_{k, \phi_k}$
\State Initialize $\theta$, $\{\phi_k\}$, and $\psi$.
\Repeat
    \State \Comment{CLIRL Phase:}
    \State Sample mini-batches from $D_{pos}$ and $D_{neg}$.
    \State Update $\theta$ and $\{\phi_k\}$ by maximizing the constraint learning objective, \cref{subsec:cl-irl-obj} via gradient ascent.
    \State \Comment{CAPO Phase:}
    \State Sample trajectories using $\pi_{\psi}$ and $T(s'|s, a, \theta)$ (sampling $\theta$).
    \State Estimate policy and constraint gradients.
    \State Update $\psi$ (e.g., trust region optimization).
\Until{convergence}
\end{algorithmic}
\end{algorithm}

\subsection{Modeling Domain Shift and Adversarial Robustness with CVaR} \label{subsec:cvar}

We address domain shift and adversarial attacks with a stochastic transition function: $T(s'|s, a, \theta)$, $\theta \in \Theta$ being a domain parameter (adversarial perturbations, topic changes, style variations) and sample $\theta$ from $P(\theta)$ during training. We also minimize the Conditional Value at Risk (CVaR) of the constraint violations:

\begin{equation}
\text{Minimize } CVaR_{\alpha}\Big(\sum_{t} \gamma^t \sum_{k=1}^K C_{k, \phi_k}(s_t, a_t)\Big)
\end{equation}

This ensures safety in worst-case scenarios.

\section{Experiment} \label{sec:experiment}

To evaluate the feasibility and adaptability of our framework, we conducted a proof-of-concept experiment in a simplified text-based navigation environment. This environment incorporates a domain shift to test the robustness of the learned constraint. The experiment's goal is to demonstrate that an instance of our framework, which we call SAfe In Language-Constraint aware Reinforcement Learning (SAIL-CaRL), can learn an initial constraint and adapt to environmental changes affecting the constraint's validity. This experiment does \emph{not} aim for state-of-the-art performance; rather, it provides a controlled demonstration.

\subsection{Environment}

We use a 5x5 grid world \cite{leike2017ai} where an agent navigates from a starting location to a goal location. States are represented textually as ``You are in room (x, y),'' where $x$ and $y$ are integer coordinates. Actions are ``go north,'' ``go south,'' ``go east,'' and ``go west.'' Transitions are deterministic: actions move the agent one cell in the corresponding direction (remaining in place if attempting to move off-grid). The agent begins at (0, 0), and the goal is at (4, 4). Initially, cell (2, 2) is a ``danger zone'' (constraint violation). After a predefined number of training epochs (\texttt{shift\_epoch} = 100), a \emph{new} danger zone is added at (3, 3), simulating say, a ``firespread'', as a domain shift. Figure~\ref{fig:gridworld} illustrates the initial environment.

\begin{figure}[h!]
\centering
\begin{tabular}{|p{0.3cm}|p{0.3cm}|p{0.3cm}|p{0.3cm}|p{0.3cm}|}
\hline
$S$ &   &   &   &   \\ \hline
  & $D_1$ &   &   &   \\ \hline
  &   & $D_2$ &   &   \\ \hline
  &   &   &   &   \\ \hline
  &   &   &   & $G$ \\ \hline
\end{tabular}
\caption{The 5x5 Safe Navigation environment. `S' denotes starting location (0, 0), `G' goal location (4, 4), and `$D_1$' initial danger zone (1,1). A second danger zone `$D_2$' is added at (2,2) after the domain shift.}
\label{fig:gridworld}
\end{figure}

\begin{figure*}[h!]
    \centering
    \includegraphics[width=\linewidth]{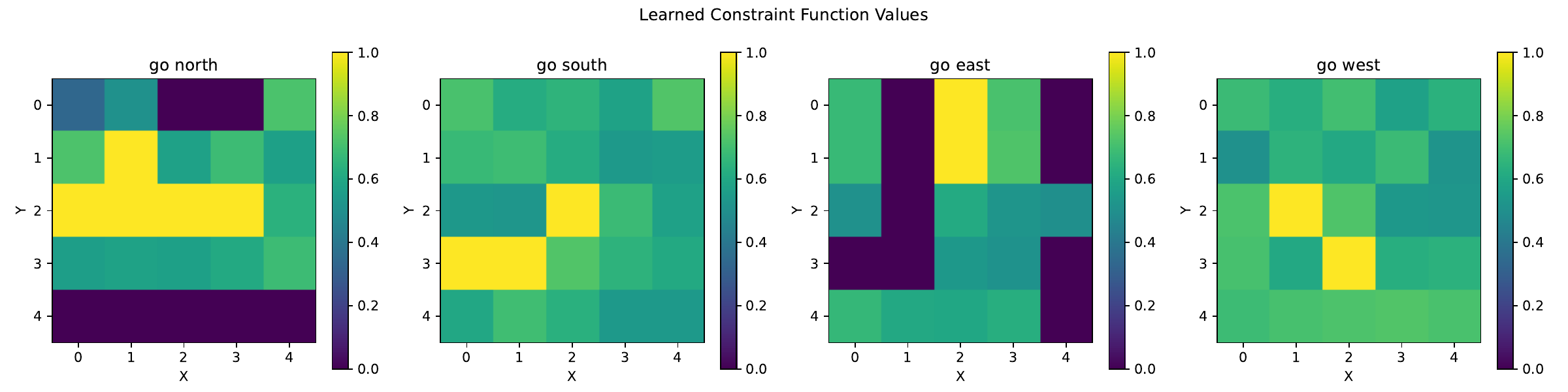}
    \caption{Learned constraint function values for SAIL-CaRL after the domain shift. Each heatmap represents an action (north, south, east, west). Brighter colors indicate a higher predicted probability of constraint violation.}
    \label{fig:learned_constraint}
\end{figure*}

Heatmaps in Figure~\ref{fig:learned_constraint} visually represent the learned constraint function after the domain shift. Critically, we observe high violation probabilities (brighter colors) for actions leading into both danger zones – (2,2) and (3,3) – from neighboring cells. For example, "go north" from (2,1) and (3,2), "go south" from (2,3) and (3,4), "go east" from (1,2) and (2,3), and "go west" from (3,2) and (4,3) all show high probabilities, as expected.  This confirms that CLIRL is learning and adapting to the new danger zone. However, the learning is imperfect. Violation probabilities are not consistently high (close to 1.0) for all danger-leading actions, and some non-dangerous actions show slightly elevated probabilities.

\subsection{Agent} We implemented a tabular version of SAIL-CaRL. A simple, predefined reward function is used: $R(s, a) = 1$ if the agent reaches the goal state, and $R(s, a) = 0$ otherwise. We focus on learning the constraint function, $C_\phi(s, a)$, a table with one entry per state-action pair. $C_\phi(s, a)$ represents the estimated probability of violating the constraint, i.e. entering a danger zone, if action $a$ is taken in state $s$. A sigmoid activation ensures a probability output. The agent's policy, $\pi_\psi(a|s)$, is also tabular, with a softmax policy: $\pi_\psi(a|s) = \exp(Q_\psi(s, a)) / \sum_{a'} \exp(Q_\psi(s, a'))$. The Q-values, parameterized by $\psi$, are learned during policy optimization.

Constraint function training uses positive ($D_{pos}$) and negative ($D_{neg}$) demonstrations. $D_{pos}$ contains trajectories reaching the goal without entering any current danger zone(s). $D_{neg}$ contains trajectories that do enter a current danger zone.
We use binary cross-entropy loss to train $C_\phi$, maximizing the likelihood of safe actions in $D_{pos}$ and unsafe actions in $D_{neg}$. The target for $C_\phi(s,a)$ is 0 (no violation) for $(s, a)$ in $D_{pos}$ and 1 (violation) for $(s, a)$ in $D_{neg}$.

Policy optimization employs a simplified policy gradient algorithm based on PPO. The objective is to maximize expected discounted return while penalizing constraint violations, based on the \emph{learned} $C_\phi$:  $J(\psi) = \mathbb{E}_{\tau \sim \pi_\psi}[\sum_t \gamma^t (R(s_t, a_t) - \beta C_\phi(s_t, a_t))]$. We use $\gamma = 0.99$ and $\beta = 0.5$. Adam is used (learning rate 0.001). Advantage normalization stabilized training.  Both CLIRL and policy training continue after the domain shift, using demonstrations generated with respect to the new danger zone configuration.

\subsection{Measurement} We compare SAIL-CaRL against two baselines: 1) ``No Constraint'': a standard policy gradient agent trained using only $R$. 2) ``Hand-coded Constraint'': a policy gradient agent with $R$ and a \emph{hand-coded} constraint function. This function assigns a violation probability of 0.99 to actions leading to any current danger zone and 0.01 otherwise. The hand-coded constraint is \emph{updated} after the domain shift, providing a strong, adaptive baseline. We use two metrics: \textit{Safe Success Rate} (percentage of episodes reaching the goal within 50 steps without entering any danger zone) and \textit{Constraint Violation Rate} (percentage of episodes entering any danger zone). We report the mean and standard deviation of both metrics over 10 independent trials, \emph{before and after} the domain shift.

\subsection{Results} \label{results}

\begin{figure*}[ht!]
    \centering
    \begin{minipage}{0.45\linewidth}
    \includegraphics[width=\linewidth]{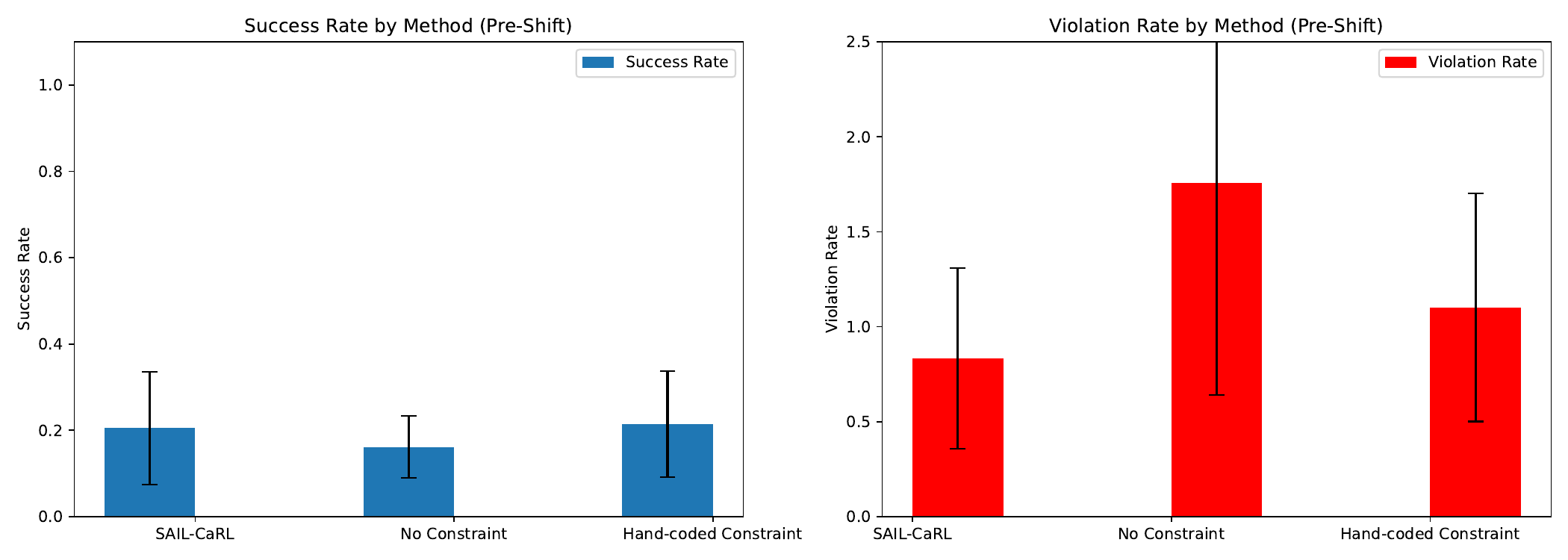}
    \caption{Agent performance prior to the domain shift. This figure presents the performance metrics for the agent; for the chart illustrating instances of zero violations, please refer to Appendix \cref{tab:addi-results}.}
    \label{fig:results_pre_shift}
    \end{minipage}\hfill
    \begin{minipage}{0.45\linewidth}
    \includegraphics[width=\linewidth]{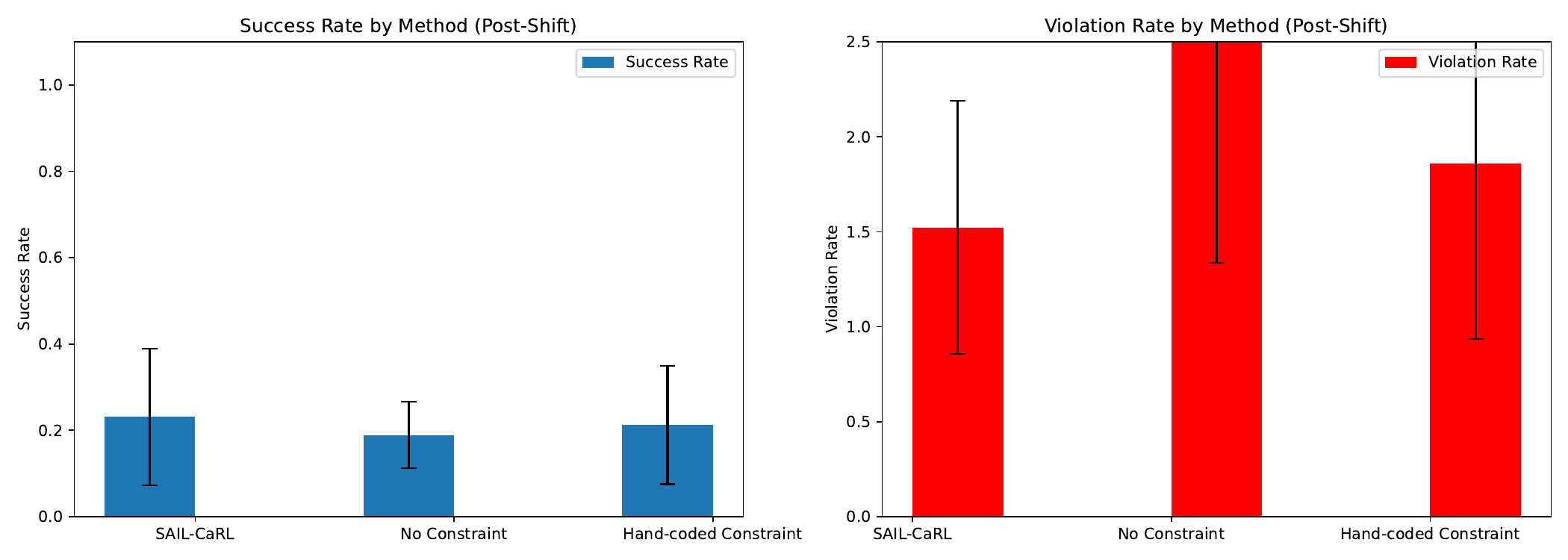}
    \caption{Agent performance following the domain shift; the agent employing SAIL-CaRL exhibits a reduced number of violations.}
    \label{fig:results_post_shift}
    \end{minipage}
\end{figure*}

Table~\ref{tab:results} presents the Safe Success Rate and Constraint Violation Rate for SAIL-CaRL and the two baselines, both \emph{before} and \emph{after} the domain shift. Figures~\ref{fig:results_pre_shift} and~\ref{fig:results_post_shift} show the pre- and post-shift results, respectively. Figure~\ref{fig:learned_constraint} shows the learned constraint function for a representative SAIL-CaRL run after the domain shift. We run a set of experiments using \emph{HuggingFace DistilBERT} tuning for around 10 hours on a single A100 GPU to demonstrate feasibility for fine-tuning and found that the LLM in gridworld \emph{violated zero constraints}.

\definecolor{highlightyellow}{HTML}{FFDE34} %
\definecolor{highlightgray}{HTML}{D3D3D3} %

\begin{table*}[!t]
\centering
\small
\begin{tabular}{lcccc}
\toprule
Method & \multicolumn{2}{c}{Pre-Shift Domain, $\theta_1$} & \multicolumn{2}{c}{Post-Shift Domain, $\theta_2$} \\
\cmidrule(lr){2-3} \cmidrule(lr){4-5}
 & Success & Violation & Success & Violation \\
\midrule
SAIL-CaRL & 0.205 ± 0.131 & 0.833 ± 0.477 & \colorbox{highlightgray}{0.231 ± 0.158} & \colorbox{highlightgray}{1.523 ± 0.665} \\
No Constraint & 0.161 ± 0.072 & 1.757 ± 1.117 & 0.189 ± 0.077 & 2.588 ± 1.251 \\
Hand-coded Constraint & 0.214 ± 0.123 & 1.102 ± 0.601 & 0.212 ± 0.137 & 1.860 ± 0.926 \\
\midrule
DistilBERT SAIL-CaRL & 0.200 ± 0.400 & \colorbox{highlightyellow}{0.000 ± 0.000} & 0.200 ± 0.400 & \colorbox{highlightyellow}{0.000 ± 0.000} \\
DistilBERT No Constraint & 0.296 ± 0.191 & 1.341 ± 0.272 & 0.289 ± 0.186 & 2.177 ± 0.687 \\
DistilBERT Hand-coded Constraint & 0.900 ± 0.300 & 0.080 ± 0.084 & 0.900 ± 0.300 & 0.036 ± 0.089 \\
\bottomrule
\end{tabular}
\caption{Experimental results on RL only and DistilBERT as base in the Safe Navigation environment, before and after the domain shift (new danger zone). Values are mean ± standard deviation over 10 trials.}
\label{tab:results}
\end{table*}

\subsection{Discussion} Before the domain shift, SAIL-CaRL's performance (Safe Success Rate: 0.205 ± 0.131, Constraint Violation Rate: 0.833 ± 0.477) is comparable to the Hand-coded Constraint baseline (Success: 0.214 ± 0.123, Violation: 1.102 ± 0.601) and slightly better than the No Constraint baseline (Success: 0.161 ± 0.072, Violation: 1.757 ± 1.117). These pre-shift results suggest that the basic CLIRL mechanism learns something about the constraint, indicated by the lower violation rate compared to No Constraint. However, the low success rates across all methods, and the high violation rate of the hand-coded constraint, highlight the challenges of this environment even before the shift. The violation rate can exceed 1 because multiple violations are possible per trajectory. After the domain shift (adding a new danger zone at (3, 3)), the performance of all methods changes. The No Constraint baseline, as expected, shows a further increase in violation rate (to 2.588 ± 1.251) and a slight increase in success rate (to 0.189 ± 0.077), being unaware of the constraints. The Hand-coded Constraint baseline's violation rate increases significantly (to 1.860 ± 0.926), with its success rate remaining similar (0.212 ± 0.137). This increase, even with a perfect constraint, likely stems from the increased difficulty of navigating with two danger zones; the simplified PPO struggles to find optimal safe paths.

\section{Open Alignment Challenges} \label{sec:open}

\paragraph{Neuro-Symbolic Integration for Reasoning}

To address reasoning limitations in purely neural systems, researchers have explored \emph{neuro-symbolic} approaches that combine sub-symbolic pattern matching with symbolic logic \citep{liu2022dualprocess,zhu2022hybridVQA}. For instance, \citet{liu2022dualprocess} integrate a neural module (System~1) for intuitive pattern recognition with a symbolic module (System~2) for precise arithmetic or logical inference. These hybrid architectures have outperformed standard neural methods on math-oriented tasks and logical NLP. Similarly, \citet{zhu2022hybridVQA} show that vision-language reasoning systems augmented with symbolic components exhibit greater robustness on out-of-distribution evaluations. %

\section{Conclusion} \label{sec:conclusion}

NLCL is a new framework that starts with learning the constraints for safe reinforcement learning augmented without augmenting human preferences. All within a CMDP, we incorporated both reward maximization and learned cost functions into the optimization objective, mitigating the shortcomings of preference learning. By leveraging positive and negative text demonstrations, our constraint-learning inverse reinforcement learning (CLIRL) procedure explicitly disentangles reward signals from safety constraints, offering safer model behaviors that can also generalize. Our experiments in a text-based navigation environment, before and after a deliberate domain shift, highlight both the promise and practical challenges of this approach. This result marks opportunity to make a synergy out of curated demonstration data, constraint architecture, and learning constraints through CLIRL in natural language to handle evolving domains.

\section{Limitations} \label{sec:limitations}

\subsection{Limitations of the natural language constraint learning framework}
While our framework offers advantages in learning constraints, it relies on the availability and quality of both positive and negative demonstration data. The framework itself does not guarantee that the learned constraints will perfectly capture all aspects of safety and alignment, nor does it address fundamental questions about whether LLMs truly understand the meaning of the constraints. The effectiveness of the framework is inherently tied to the data used to train it, and biases or omissions in the data could lead to unintended consequences. As such, there is ongoing debate on whether large language models (LLMs) genuinely understand language or merely learn statistical patterns from data \citep{bender2020climbing,vanDijk2023foundation}. \citet{bender2020climbing} argue that systems trained solely on form cannot fully capture meaning, cautioning against conflating fluent output with semantic comprehension. Conversely, \citet{vanDijk2023foundation} contend that LLMs may exhibit functional competence in context, even through mechanisms different from human cognition. This pragmatic perspective suggests that attributing ``understanding'' can be useful for predicting model behavior, while acknowledging that form-based learning alone may not equate to natural language semantic grounding \cite{richens2024robust}.

\bibliography{anthology,custom}

\appendix

\section{Constraint Learning Max Inverse Reinforcement Learning (CLIRL)} \label{subsec:cl-irl-obj}

As was discussed in \cref{subsec:cl-irl}, the core innovation of our framework is Constraint Learning Inverse Reinforcment Learning (CLIRL), changing IRL to learn rewards and constraints. We use positive demonstrations, $D_{pos}$ ($\{\tau_i^+\}$ of desirable behavior), and negative demonstrations, $D_{neg}$ ($\{\tau_j^-\}$ of undesirable behavior), and details of the objective is detailed here.

Reward and constraints are parameterized as $R_{\theta}(s, a)$ and $C_{k, \phi_k}(s, a)$, with learnable parameters $\theta$ and $\phi_k$. CLIRL objective adapts Maximum Causal Entropy IRL. It maximizes positive demonstration likelihood while minimizing negative demonstration likelihood under a combined reward and cost model:

\begin{equation}
\label{eq:cairl_objective}
\begin{aligned}
\mathcal{L}(\theta, \{\phi_k\}) &= \sum_{\tau^+ \in D_{pos}} \log P_{\theta}(\tau^+)  \\
&- \sum_{\tau^- \in D_{neg}} \log P_{\theta, \{\phi_k\}}(\tau^-)  \\
&- \lambda \sum_{k=1}^K \Big( \mathbb{E}_{\pi_{\theta, \{\phi_k\}}} \Big[ \sum_{t=0}^{\infty} \gamma^t C_{k, \phi_k}(s_t, a_t) \Big] \\
&\quad - H_k \Big)^2
\end{aligned}
\end{equation}

Where:

\begin{equation}
\begin{aligned}
P_{\theta}(\tau^+) &\propto \exp\Big(\sum_{t} \gamma^t R_{\theta}(s_t^+, a_t^+)\Big) \\
P_{\theta, \{\phi_k\}}(\tau^-) &\propto \exp\Big(\sum_{t} \gamma^t \big[R_{\theta}(s_t^-, a_t^-) \\
&\quad - \sum_{k=1}^K \alpha_k C_{k, \phi_k}(s_t^-, a_t^-)\big]\Big)
\end{aligned}
\end{equation}

The final term penalizes constraint violations ($\lambda$ controls strength).

$\pi_{\theta, \{\phi_k\}}$ is the policy from $R_\theta$ and $C_{k, \phi_k}$.

As described in the main text, after policy learning our method discovers safety constraints, not manual specifications. Negative demonstrations are key. For example, in a dialogue setting, a negative demonstration might be a conversation turn where the LLM generates a toxic response, reveals private information, or provides a factually incorrect answer.  In a text-based game, a negative demonstration could be a sequence of actions that leads to a game-over state due to violating a safety rule (e.g., drinking a poisonous potion or walking into a bottomless pit).

\section{Perspectives on Language Model Understanding: Form vs Meaning in Language Models}

The success of large pre-trained language models (LLMs) on many NLP tasks has sparked considerable discussion, and often hype, about whether these models truly understand language or merely learn superficial patterns.  While some popular accounts have suggested LLMs capture "meaning," a more nuanced academic debate is ongoing since. \citet{bender-koller-2020-climbing} forcefully argue that a system trained only on linguistic form (i.e., text) has no \emph{a priori} way to learn meaning, since meaning ultimately derives from grounding in the world and communicative intent. This perspective suggests that no matter how much text a model consumes, it lacks natural language understanding. On the other hand, subsequent work has shown that purely form-based learners can acquire a surprising amount of relational and factual knowledge from text alone. Ever since the promise of \citet{petroni-etal-2019-language} BERT that contains relational knowledge showed it can answer fill-in-the-blank queries at a level competitive with systems that explicitly leverage curated knowledge bases. Models like BERT and its successors also exhibit a strong ability to recall factual information without any fine-tuning, effectively functioning as unsupervised open-domain QA systems \citep{roberts-etal-2020-much}. Such findings indicate that some aspects of what we might consider knowledge, or even precursors to meaning, can be learned from form alone, challenging the strict view that form and meaning are entirely disjoint. This tension between the "form is sufficient" perspective and the need for grounding remains a central open question in NLP. While distributional semantics posits that word meaning can be derived from usage patterns, skeptics maintain that true understanding requires more than just statistical correlations extracted from text. The question of how to build LLMs that are both knowledgeable and safe is closely related to this debate.  If a model lacks a grounded understanding of the world, can it reliably avoid generating harmful or misleading content?  This motivates the development of techniques like Safe Reinforcement Learning (Safe RL). The field continues to explore how far we can push form-based learning before hitting a ceiling where additional grounding or structured knowledge becomes necessary. Our work on Safe RL in text-based environments contributes to this exploration. We conclude that text-based environments serve as a controlled yet expressive sandbox for developing safe, interpretable, and generalizable language agents, offering a way to test the limits of form-based learning while simultaneously addressing crucial safety concerns, and thereby indirectly informing the debate on the relationship between form, meaning, and grounding in LLMs.

\section{Text-Based Environments as a Structured Evaluation Ground}

Text interactive environments are valuable testbeds for studying generalization and safety in RL-based NLP. These environments present partially observable, language-mediated worlds where agents read descriptions and execute text commands \cite{osborne2022survey}. In addition, it's important to point out that they provide a controlled yet realistic proxy for real-world  language tasks: the agent expriences a variety of scenarios described in natural language, but within a sandbox where outcomes and rewards are well-defined. This makes it easier to evaluate whether an agent truly understands and generalizes the task. In fact, text games are considered a safe and data-efficient platform for RL research, “mimic(king) language found in real-world scenarios” while avoiding physical risks. Rewards in text games are valuable for safety research precisely because they make the reward-goal relationship explicit through language. When a quest states 'Find the treasure hidden in the kitchen' and provides points for completing this task, we can directly analyze whether the agent's understanding matches the stated goal. This linguistic specification of objectives allows us to detect misalignment between the reward signal and intended behavior by comparing the agent's actions against the explicit textual instructions. As such, rewards in these games (points, quest completion) are typically simple to specify and tightly correlated with the goal, reducing ambiguity in feedback, and that makes an agent’s tendency to exploit reward loopholes or generalize incorrectly that can be readily observed and analyzed before presumed readiness for generalization and deploying similar techniques in open-ended NLP tasks.
Another advantage of text environments is the scope for integrating structured knowledge and hierarchical reasoning, which can be critical for both generalization and safety. Researchers have leveraged knowledge graphs to represent the game state, where entities, locations, and their relations discovered through exploration are stored in a graph memory \cite{ammanabrolu2018playing}. This approach helps to manage the combinatorial action space by pruning irrelevant actions and focusing the agent’s decisions on causally relevant factors \cite{ammanabrolu2020graph, adhikari2020learning}. Similarly, agents can update an explicit graph of the world as they explore, gradually improving on an ever more accurate representation of the environment that improves long-term planning \cite{chhikara2023knowledge}. On top of such representations, hierarchical RL techniques have been applied: a high-level policy breaks down the overall goal into sub-goals or subtasks (often readable in text form), and a low-level policy is charged with executing each subtask \cite{xu2021generalization}. \citet{xu2021generalization} implement this by having a meta-controller choose textual sub-goals based on the knowledge graph state, and a subordinate controller then pursues each sub-goal, leading to improved generalization across games of varying difficulty. This kind of hierarchy mirrors how humans approach complex quests (first get the key, then open the door, then enter the treasure room), and it can prevent the agent from getting sidetracked by irrelevant behaviors, thereby mitigating goal misgeneralization within the game’s context.
Moreover, text games often come with natural language instructions or narratives that specify the desired outcomes (“find the treasure hidden in the kitchen”). Harnessing such guidance is an active research area. While one might expect an RL agent to naturally follow in-game instructions, state-of-the-art agents have been found to largely ignore them and performing no better with instructions present than absent. This indicates that without special design, agents don’t inherently understand or utilize textual guidance \cite{huang2022language}. To address this, instruction-guided architectures translate language instructions into structured objectives. For instance, recent work encodes game instructions as Linear Temporal Logic (LTL) formulas that the agent can explicitly plan over. In incorporating a formal representation of the instructions into the reward and policy (e.g. giving intermediate rewards for satisfying parts of an LTL goal), agents achieved significantly better task completion rates in over 500 TextWorld games \cite{tuli2022learning}. This demonstrates that text-based environments as a safe harbor not only allow us to evaluate generalization and safety in a controlled manner, but also to experiment with injecting high-level knowledge (via graphs, hierarchies, or instructions) to guide learning. In our context, these environments will serve as a proving ground for the agent’s ability to generalize safely as they provide a repeatable way to test if new reward functions and constraints truly prevent misbehavior under varied conditions.

\subsection{Additional Results} \label{add-results}

Table~\ref{tab:addi-results} presents the Safe Success Rate and Constraint Violation Rate for SAIL-CaRL and the two baselines, both \emph{before} and \emph{after} the domain shift. Figures~\ref{fig:results_pre_shift} and~\ref{fig:results_post_shift} show the pre- and post-shift results, respectively. Figure~\ref{fig:learned_constraint} shows the learned constraint function for a representative SAIL-CaRL run after the domain shift. We run a set of experiments using \emph{HuggingFace DistilBERT} tuning for around 10 hours on a single GPU to demonstrate feasibility for fine-tuning and found that the LLM in gridworld \emph{violated zero constraints}.

\definecolor{highlightyellow}{HTML}{FFDE34} %
\definecolor{highlightgray}{HTML}{D3D3D3} %

\begin{table*}[!t]
\centering
\small
\begin{tabular}{lcccc}
\toprule
Method & \multicolumn{2}{c}{Pre-Shift Domain, $\theta_1$} & \multicolumn{2}{c}{Post-Shift Domain, $\theta_2$} \\
\cmidrule(lr){2-3} \cmidrule(lr){4-5}
 & Success & Violation & Success & Violation \\
\midrule
SAIL-CaRL & 0.205 ± 0.131 & 0.833 ± 0.477 & \colorbox{highlightgray}{0.231 ± 0.158} & \colorbox{highlightgray}{1.523 ± 0.665} \\
No Constraint & 0.161 ± 0.072 & 1.757 ± 1.117 & 0.189 ± 0.077 & 2.588 ± 1.251 \\
Hand-coded Constraint & 0.214 ± 0.123 & 1.102 ± 0.601 & 0.212 ± 0.137 & 1.860 ± 0.926 \\
\midrule
DistilBERT SAIL-CaRL & 0.200 ± 0.400 & \colorbox{highlightyellow}{0.000 ± 0.000} & 0.200 ± 0.400 & \colorbox{highlightyellow}{0.000 ± 0.000} \\
DistilBERT No Constraint & 0.296 ± 0.191 & 1.341 ± 0.272 & 0.289 ± 0.186 & 2.177 ± 0.687 \\
DistilBERT Hand-coded Constraint & 0.900 ± 0.300 & 0.080 ± 0.084 & 0.900 ± 0.300 & 0.036 ± 0.089 \\
\bottomrule
\end{tabular}
\caption{Experimental results on RL only and DistilBERT as base in the Safe Navigation environment, before and after the domain shift (new danger zone). Values are mean ± standard deviation over 10 trials.}
\label{tab:addi-results}
\end{table*}

\begin{figure*}[ht!]
    \centering
    \includegraphics[width=\linewidth]{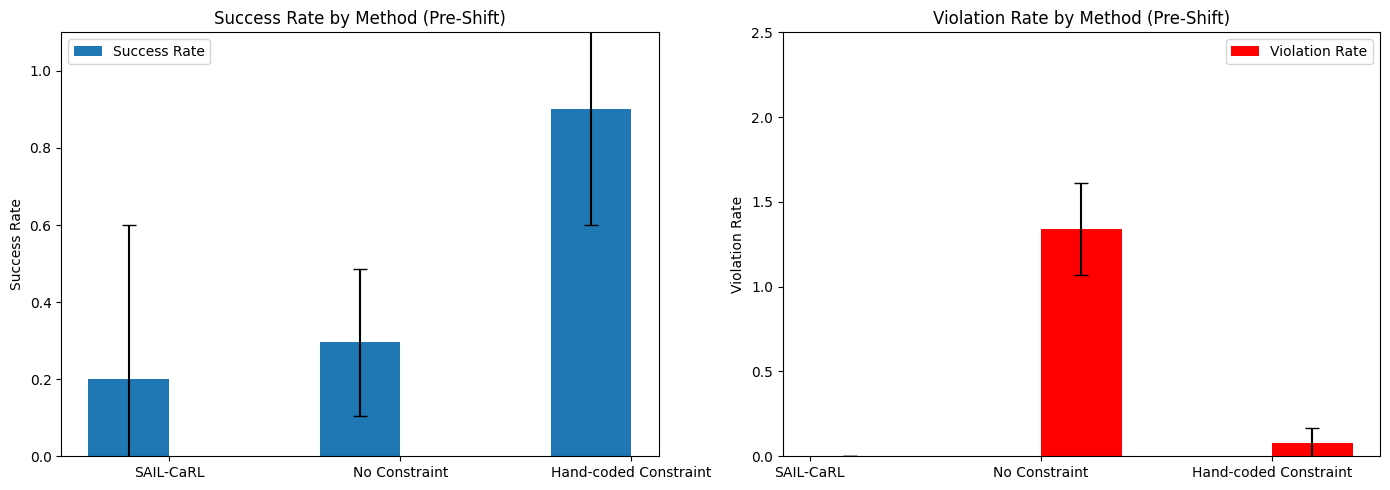}
    \caption{CMDP + DistilBERT results chart as base with zero violations.}
    \label{fig:results_pre_shift_bert}
    \includegraphics[width=\linewidth]{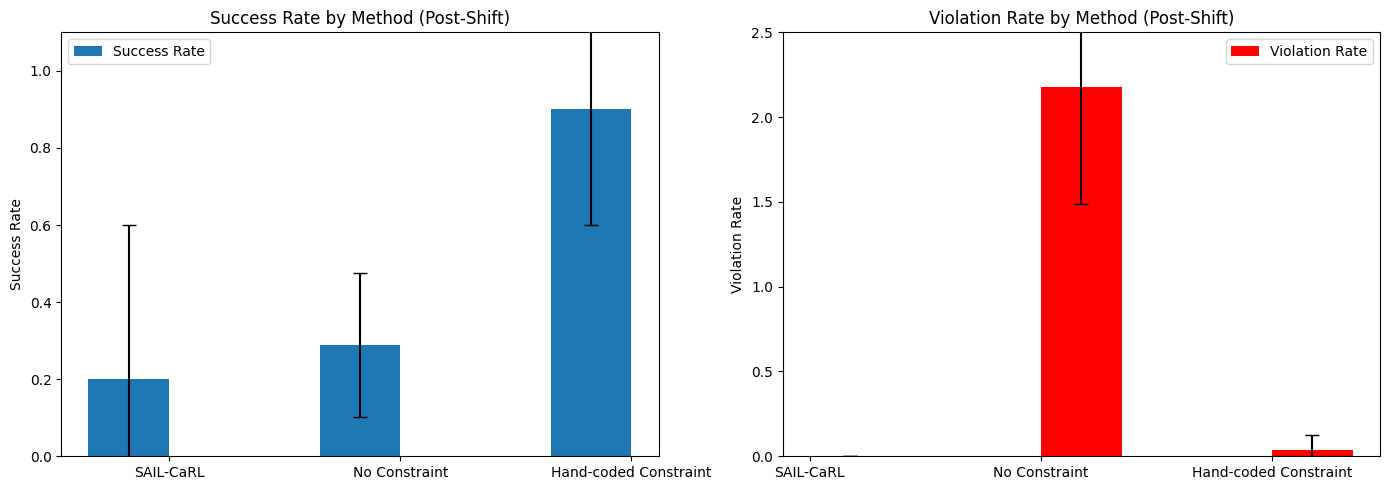}
    \caption{CMDP + DistilBERT results chart as base with zero violations.}
    \label{fig:results_post_shift_bert}
\end{figure*}

\end{document}